\documentclass[conference]{IEEEtran}         
\IEEEoverridecommandlockouts
\usepackage[utf8]{inputenc}
\usepackage[T1]{fontenc}
\usepackage{CJKutf8}

\usepackage{cite}
\usepackage{amsmath,amssymb,amsfonts}
\usepackage{algorithmic}
\usepackage{graphicx}
\usepackage{textcomp}
\usepackage{xcolor}
\usepackage{xurl}
\usepackage{cleveref}
\usepackage{tabularx}
\usepackage{subcaption}
\usepackage{multirow}
\usepackage{booktabs, multirow, siunitx, pifont}
\newcommand{\cmark}{\ding{51}} 
\usepackage{dblfloatfix}

\usepackage[linesnumbered,ruled,vlined,commentsnumbered]{algorithm2e}

\SetCommentSty{mycommfont}

\def\appendixautorefname~#1\null{~#1 \null}
\DeclareMathSymbol{\mhyph}{\mathalpha}{operators}{`-}
\newcommand{\xmark}{\ding{55}}

\def\equationautorefname~#1\null{Eq.~(#1)\null}

\def\BibTeX{{\rm B\kern-.05em{\sc i\kern-.025em b}\kern-.08em
    T\kern-.1667em\lower.7ex\hbox{E}\kern-.125emX}}
    
\begin{document}

\title{Agent Fine-tuning through Distillation for Domain-specific LLMs in Microdomains \\

}

\author{\IEEEauthorblockN{Yawen Xue}
\IEEEauthorblockA{\textit{Research and Development Group} \\
\textit{Hitachi, Ltd.}\\
Tokyo, Japan \\
yawen.xue.wn@hitachi.com}
\and
\IEEEauthorblockN{Masaya Tsunokake}
\IEEEauthorblockA{\textit{Research and Development Group} \\
\textit{Hitachi, Ltd.}\\
Tokyo, Japan \\
masaya.tsunokake.qu@hitachi.com}
\and
\IEEEauthorblockN{Yuta Koreeda}
\IEEEauthorblockA{\textit{Research and Development Group} \\
\textit{Hitachi, Ltd.}\\
Tokyo, Japan \\
yuta.koreeda.pb@hitachi.com}
\and
\IEEEauthorblockN{Ekant Muljibhai Amin}
\IEEEauthorblockA{\textit{Research and Development Group} \\
\textit{Hitachi, Ltd.}\\
Tokyo, Japan \\
ekant.amin.mu@hitachi.com}
\and
\IEEEauthorblockN{Takashi Sumiyoshi}
\IEEEauthorblockA{\textit{Research and Development Group} \\
\textit{Hitachi, Ltd.}\\
Tokyo, Japan \\
takashi.sumiyoshi.bf@hitachi.com}
\and
\IEEEauthorblockN{Yasuhiro Sogawa}
\IEEEauthorblockA{\textit{Research and Development Group} \\
\textit{Hitachi, Ltd.}\\
Tokyo, Japan \\
yasuhiro.sogawa.tp@hitachi.com}
\and
}

\maketitle

\begin{abstract}
Agentic large language models (LLMs) have become prominent for autonomously interacting with external environments and performing multi-step reasoning tasks. Most approaches leverage these capabilities via in-context learning with few-shot prompts, but this often results in lengthy inputs and higher computational costs. Agent fine-tuning offers an alternative by enabling LLMs to internalize procedural reasoning and domain-specific
knowledge through training on relevant data and demonstration trajectories. While prior studies have focused on general domains, their effectiveness in specialized technical microdomains remains unclear. This paper explores agent fine-tuning for domain adaptation within Hitachi's JP1 middleware, a microdomain for specialized IT operations. We fine-tuned LLMs using JP1-specific datasets derived from domain manuals and distilled reasoning trajectories generated by LLMs themselves, enhancing decision making accuracy and search efficiency. During inference, we used an agentic prompt with retrieval-augmented generation
and introduced a context-answer extractor to improve information relevance. On JP1 certification exam questions, our
method achieved a 14\% performance improvement over the base model, demonstrating the potential of agent fine-tuning for
domain-specific reasoning in complex microdomains.

\end{abstract}

\begin{IEEEkeywords}
Large language models, agent fine-tuning, microdomains, domain adaptation, retrieval-augmented generation, JP1 middleware, structured reasoning.
\end{IEEEkeywords}

\section{Introduction}
Large language models (LLMs) such as GPT-4~\cite{openai2023gpt4, openai2023blog}, Gemini~\cite{gemini2023}, Llama~\cite{vavekanand2024llama,llama3modelcard}, Qwen~\cite{qwen2023}, and DeepSeek~\cite{deepseek2023} have shown strong generalization across diverse language tasks due to training on large-scale text corpora~\cite{xu2025survey}. Recently, \textbf{agentic LLMs}~\cite{qiu2024llm,ferrag2025llm} have emerged, combining language understanding with the ability to interact autonomously with external tools and environments, enabling multi-step reasoning and goal-directed actions.

Many prior studies have explored the use of \textbf{few-shot prompting} \cite{sprague2024cot, wei2022chain, yao2023react} to enhance LLMs' agentic capabilities. These approaches provide models with in-context examples to guide their reasoning and decision-making without requiring additional fine-tuning. While effective to some extent, few-shot prompting has limitations in consistency, robustness, and adaptability. To overcome these challenges, recent research has explored \textbf{agent fine-tuning} \cite{fireact2024, qiao2024autoact, peng2024improving, chen2024agent, yin2023lumos, yin2023agent} on general domain, where LLMs are explicitly trained to act as agents rather than relying solely on prompting. FireAct \cite{fireact2024} fine-tunes LLMs on GPT-4-generated agentic trajectories across diverse QA tasks with a Google-search tool, outperforming prompt-only agents in reasoning and tool use. 
AUTOACT \cite{ qiao2024autoact} further introduces a fully automatic agentic learning framework that synthesizes planning trajectories without human or strong model assistance. 
Previous research on agent fine-tuning  \cite{fireact2024, qiao2024autoact, peng2024improving, chen2024agent, yin2023lumos, yin2023agent} has concentrated primarily on general domains, whereas its usefulness in
specialized technical microdomain remains unclear.

In this paper, we explore agent fine-tuning for domain adaptation within Hitachi’s JP1 middleware, a highly specialized IT micro-domain. Our approach involves continual pre-training (CPT) and supervised fine-tuning (SFT) of LLMs on JP1-specific data, which includes comprehensive manuals and distilled agent trajectories generated by the LLMs. Unlike the previous study ~\cite{fireact2024}, which utilized a Google Search API to retrieve simple answers, our work focuses on domain-specific data that cannot be effectively addressed through general-purpose search engines. To tackle this challenge, we employ retrieval-augmented generation (RAG) \cite{lewis2020retrieval, yu2024evaluation} for JP1 tasks. Additionally, to extract key information from lengthy retrieved contexts, we introduce a contextual answer extractor that identifies and highlights salient details within the retrieved passages.
We evaluate our method on multiple-choice questions (MCQs) from the JP1 certification exam, guiding problem solving with agentic trajectories. Experiments show that our pipeline improves mean accuracy by 14\% percentage points over the base model and even outperforms GPT-4 (gpt-4-0618) \cite{achiam2023gpt} on the JP1 Professional and Consultant certification benchmark. These results underscore the value of agent fine-tuning for domain-specific reasoning in complex microdomains.

\section{JP1-Oriented Agentic LLM}
\label{sec:jp1_agent}

\begin{figure}[t]
    \centering
    \includegraphics[width=\columnwidth]{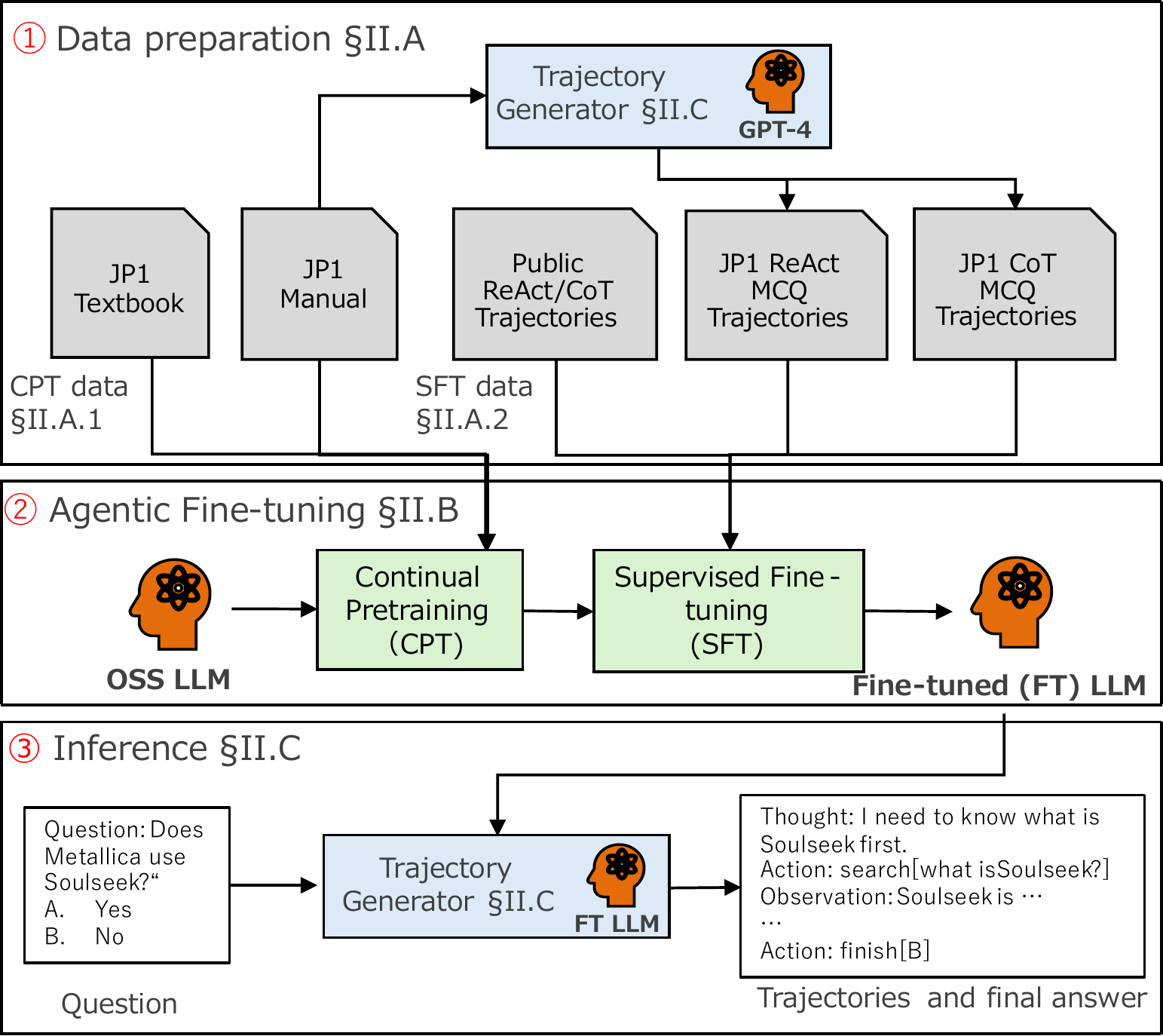}
    \caption{Pipeline of Language Agent Fine-Tuning}
    \label{fig:agent_ft_pipeline}
\end{figure}

This section presents the fine-tuning pipeline for developing Hitachi's JP1-specialized language agent, leveraging an agent fine-tuning framework. The process consists of three key stages: (1) data preparation, (2) agent fine-tuning, which includes the CPT and SFT, and (3) inference. \Cref{fig:agent_ft_pipeline} illustrates this pipeline.

The initial phase focuses on gathering pre-training text data from JP1 manuals and textbooks, forming a robust foundation for the task. Additionally, an agent trajectory dataset is curated, comprising both publicly available and domain-specific data. These trajectories are meticulously generated using advanced methodologies such as reasoning trajectories of multiple iternations (ReAct) and only one iterations (CoT), leveraging the capabilities of GPT-4 for enhanced precision and contextual understanding. 

In the second stage, known as agent fine-tuning, we begin by training a domain-specific LLM that has already undergone the CPT process using the pre-training text data. Following this, SFT is applied to the domain-specific LLM, utilizing agent-oriented datasets. This process refines the model further, resulting in a highly specialized LLM that is better equipped for reasoning and decision-making tasks within the target domain.

During the inference phase, we focused on generating agent trajectories through a process referred to as the trajectory generator, as illustrated in \Cref{fig:react_procedure}. The fine-tuned LLM processes user queries and dynamically selects the optimal reasoning approach, employing either CoT or ReAct trajectories. To further enhance reasoning capabilities, a retrieval tool is incorporated, enabling the agent to effectively access and utilize external knowledge. In this paper, we introduce a novel context answer extractor to improve retrieval efficiency. This module leverages an LLM to distill relevant information from lengthy retrieved contexts, ensuring that only the most pertinent knowledge is retained. By utilizing structured agent trajectories, the model consistently produces responses that are well-reasoned, contextually appropriate, and highly informative.

\subsection{Construction of CPT data}
To ensure reproducibility, we collected data from JP1 manuals and textbooks using systematic methods. JP1 manuals were accessed online\footnote{\url{https://itpfdoc.hitachi.co.jp/Pages/document_list/manuals/jp1v12.html}} and prioritized in HTML format due to its structured nature and ease of text extraction; PDFs were used when HTML versions were unavailable. JP1 textbooks, provided in Microsoft Word (\texttt{.doc}) format, were processed using the \texttt{python-docx} library\footnote{\url{https://github.com/python-openxml/python-docx}} to extract text from paragraphs and tables, with tables converted into HTML format for compatibility with LLMs\cite{sui2024table}. To ensure data integrity, we confirmed no test data leakage by identifying duplicate substrings\cite{lee-etal-2022-deduplicating} and manually reviewing examples. Summary statistics of the dataset are provided in \Cref{tab:data_statistics}.
\textcolor{red}{\begin{table*}[t]
    \centering
    \caption{List of datasets used for LLM training}\label{tab:data_statistics}
    \fontsize{8pt}{9pt}\selectfont
    \setlength{\tabcolsep}{6pt}
    \renewcommand{\arraystretch}{1.1}
    \begin{tabular}{lllp{7cm}rr}
        \toprule
        \multicolumn{2}{l}{\textbf{Data}} & \textbf{Use} & \textbf{Description} & \textbf{Size(MB)} & \textbf{\#Doc./Sample} \\
        \midrule
        JP1 Manual & (HTML) & CPT & Japanese text extracted from online JP1 HTML manuals & 83.7 & 42,357 \\
               & (PDF) & CPT & Japanese text extracted from online JP1 PDF manuals & 148.6 & 337 \\
        \multicolumn{2}{l}{JP1 Textbook} & CPT & Japanese text extracted from proprietary JP1 textbooks & 3.3 & 70 \\
        \multicolumn{2}{l}{JP1 ReAct MCQ trajectories} & SFT & Using GPT-4, we generated ReAct trajectories for the MCQ dataset derived from the JP1 manual. & 22.6  & 8,044  \\
        \multicolumn{2}{l}{JP1 CoT MCQ trajectories} & SFT & CoT trajectories of MCQ dataset created from JP1 manual with GPT-4 & 4.2 & 3,404 \\
        \multicolumn{2}{l}{Public ReAct and CoT trajecotires} & SFT & Public dataset containing ReAct and CoT trajectories from multiple tasks in both English and Japanese. & 4.2 & 4,142 \\
        \bottomrule
    \end{tabular}
\end{table*}
}
\subsection{Construction of SFT Data}\label{sec:dataset-sft}
To align the model with JP1-specific tasks, we construct SFT data using two categories of agent trajectories: JP1 domain data and public domain data. These datasets are used to refine the model's problem-solving and reasoning capabilities through structured ReAct and CoT trajectories. 

\paragraph{JP1 Domain Data}
JP1 domain-specific ``thought, action, observation'' trajectories  are generated using GPT-4 (gpt-4-0618), leveraging JP1 MCQs distilled from JP1 manuals to ensure strong domain relevance. The prompt used for distillation follows the same format as the FireAct project~\cite{fireact2024}, with the addition of the instruction ``Please answer in Japanese'' to ensure responses are generated in Japanese. These trajectories are based on the ReAct framework, which combines reasoning and action to dynamically process queries. A special case of this framework is CoT, where the iteration is limited to a single step, enabling step-by-step answer derivation. By utilizing both ReAct and its streamlined CoT variant, the model is equipped with structured reasoning capabilities tailored specifically to JP1-related tasks, ensuring accurate and context-aware performance within the domain. 





\paragraph{Public Domain Data}
To enhance generalization and robustness, we integrate publicly available agent trajectories from the FireAct project\footnote{\url{https://github.com/anchen1011/FireAct/blob/main/data/finetune/alpaca_format/multitask_multimethod.json}}. This dataset includes both ReAct and CoT data across a variety of tasks. To create a bilingual training corpus, we further translate the dataset into Japanese using DeepL Pro, allowing the model to handle multilingual scenarios effectively. An illustrative example of the public domain data is shown in the inference pipeline in \Cref{fig:agent_ft_pipeline}.

\subsection{Agent Fine-tuning}
Before SFT, we conduct CPT to enhance the model’s understanding of JP1-related concepts and domain-specific terminology. This phase focuses on training the LLM using next-word prediction on a corpus built from JP1-specific documents, including the JP1 Manual and JP1 Textbook. By pretraining on this domain-specific corpus, the model acquires JP1-related linguistic patterns and technical knowledge, serving as a foundation for the subsequent agent fine-tuning stage.

The agent fine-tuning is conducted as SFT with the generated SFT datasets described in \Cref{sec:dataset-sft}. During SFT, the model is trained on the agent trajectories consisting of structured prompts and corresponding high-quality responses that reflect reasoning patterns. Through this process, the fine-tuned LLM becomes more reliable in answering JP1-related queries while maintaining structured reasoning capabilities. We use the same prompt as FireAct~\cite{fireact2024} for SFT.

\subsection{Inference}

\begin{figure*}[t]
    \centering
    \includegraphics[width=\textwidth]{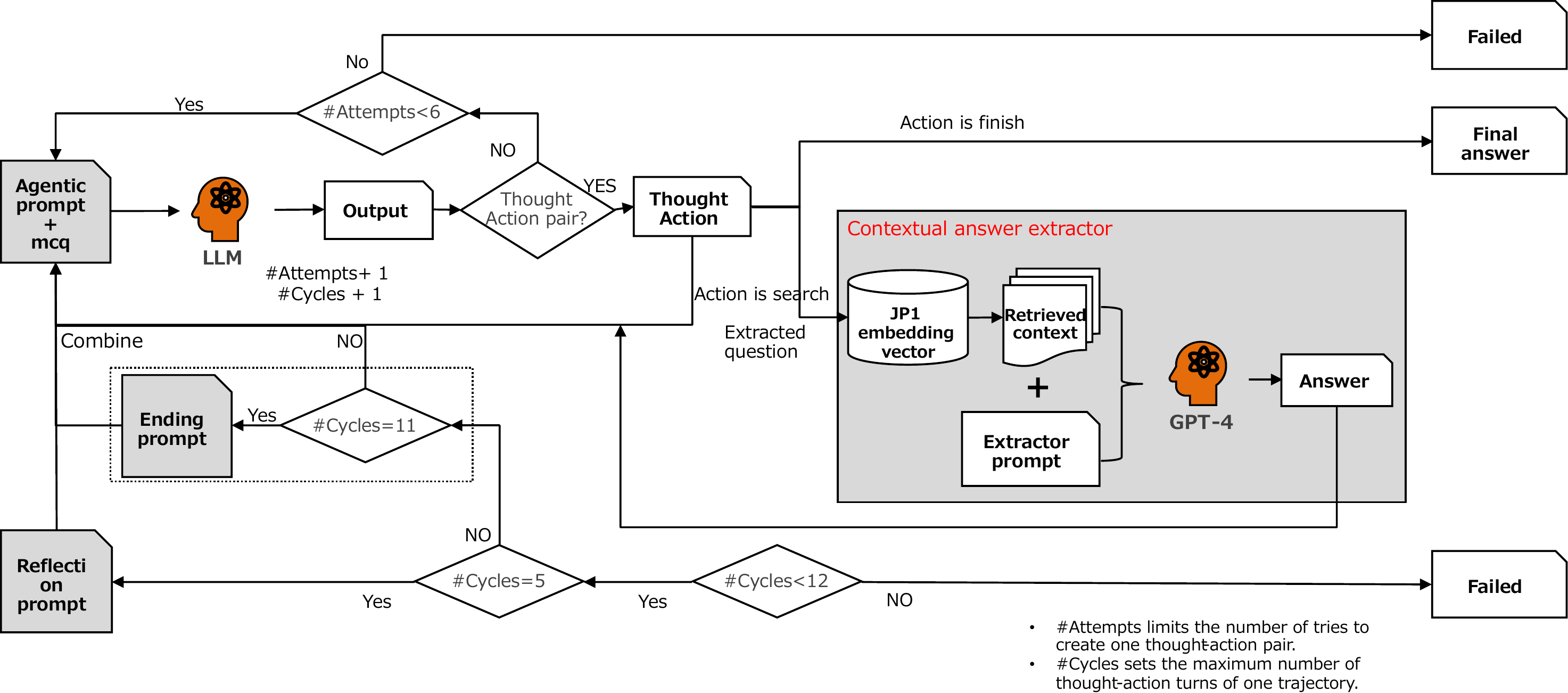}
    \caption{Illustration of the process for generating agent trajectories during inference, referred to as the Trajectory Generator.}
    \label{fig:react_procedure}
\end{figure*}

To evaluate the effectiveness of our fine-tuned language agent, we employ two inference methods: standard prompting and agentic prompting. Each method assesses the model's ability to answer MCQs while leveraging different reasoning strategies.

\subsubsection{Standard Prompting}
In this approach, the model receives an MCQ as input and directly generates the most likely answer choice. This method relies solely on the model's pre-trained and fine-tuned knowledge without explicit reasoning steps. The input to the model is MCQ and output the predicted answer choice (one of A, B, C, or D).
This approach provides a direct answer without intermediate reasoning, making it computationally efficient while leveraging the model's fine-tuned knowledge.

\subsubsection{Agentic Prompting}\label{sec:react prompting}
To further enhance reasoning capabilities, we employ agentic prompting, utilizing the same prompt framework as in the agent fine-tuning phase. This consistency ensures alignment between training and inference, allowing the model to effectively leverage domain-specific reasoning patterns. The inference process proceeds through several key steps, as illustrated in \Cref{fig:react_procedure}, to guide the generation of well-structured and contextually appropriate responses. In particular, the steps involving iterative reasoning with thought-action pairs and the contextual answer extractor are repeated multiple times until reaching the predefined maximum number of attempts or cycles.

\paragraph{Step 1: Initial Prompting}
The inference process begins with a carefully crafted structured prompt that guides the model to generate responses using step-by-step reasoning while ensuring the final answer is selected as a single choice from $<$A, B, C, D$>$. This prompt integrates agentic prompting with the MCQ format, providing clear instructions for the reasoning process. By combining these elements, the structured prompt ensures consistency and enhances the accuracy of the model's output.

To further refine the inference process, we initialize two key parameters:

    \begin{itemize}
        \item \#Attempts: This parameter controls the number of attempts the system makes to generate a valid thought-action pair during each reasoning cycle. If no valid pair is found within the defined limit, the process terminates without producing an answer.
        \item \#Cycles: This counter tracks the total number of reasoning cycles performed before arriving at a final answer. If the maximum limit for cycles is exceeded, the system halts and does not generate a response.
    \end{itemize}

These parameters bring a structured approach to the reasoning process, maintaining control over the iterations while ensuring the system adheres to defined constraints for generating accurate and contextually appropriate answers.

\paragraph{Step 2: Iterative Reasoning with Thought-Action Pairs}
After processing the input, the LLM generates an output in the form of thought-action pairs. The system evaluates the response as follows: If no thought-action pair is detected, the iteration counter \#Attempt is incremented by 1, and the agentic prompt is applied again. Since the temperature of the LLM is set to a high value, it is assumed that different results will be produced in subsequent iterations. On the other hand, if a thought-action pair is detected, the system proceeds to execute the corresponding action.

\paragraph{Step 3: Contextual Answer Extractor}
If the action derived from the last step's thought-action pair is \texttt{search}, the system proceeds to the retrieval module to gather relevant information for the query.
One limitation of traditional methods, such as those purely relying on RAG, is that they often retrieve lengthy contexts that may contain both relevant and extraneous information. LLMs frequently struggle to efficiently extract useful details from such extensive contexts, which can lead to diluted or imprecise responses.
To address this issue, we propose the \textit{Contextual Answer Extractor}, a module designed to selectively extract and refine the most relevant information from the retrieved context before passing it to the reasoning module. When an action involves a search query, the system activates the Contextual Answer Extractor, which retrieves information from the JP1 Embedding Vector Storage. This approach enhances retrieval-augmented reasoning by improving accuracy and incorporating domain-specific knowledge while mitigating the challenges of handling excessively long contexts.
The process includes: retrieving relevant context based on the extracted query, merging the retrieved context, and using GPT-4 to generate a refined answer.

\paragraph{Step 4: Iterative Reflection and Answer Finalization}
After generating an initial response, the system evaluates whether further refinement is necessary. If the model reaches five iterations without producing a satisfactory answer, the Reflection Prompt is issued, instructing the model to analyze previous failures, identify patterns, and propose an improved reasoning and action plan. If an answer is still not finalized, the system continues with further iterations under the same evaluation framework. Upon reaching the final iteration, the Final Iteration Prompt is issued, requiring the model to generate a definitive answer using \texttt{Action: finish[x]}, where $x$ is one choice from \texttt{<A, B, C, D>}, ensuring consistency with the reasoning established in prior steps.

\section{Experiment Setup}
\subsection{Model training details}
We conduct CPT with standard autoregressive language modeling objective using JP1 manual on \textbf{Llama3.1--70B-Japanese-Instruct}\cite{cyberagent-llama-3.1-70b-japanese-instruct-2407}, which had previously undergone continual pretraining in Japanese on the base model Llama3.1-70B. We use the AdamW optimizer \cite{DBLP:journals/corr/abs-1711-05101} with hyperparameters set to $\beta_1=0.9$, $\beta_2=0.95$, and weight decay to 0.1.
The scheduler is set to constant, with the learning rate configured at $1.6 \times 10^{-5}$. 
We train models for 3 epochs, using a context length of 2,048 and an effective mini-batch size\footnote{24 GPUs, each with mini-batch size of one and gradient accumulation steps of five ($24 \times 1 \times 5 = 160$). Our training was affected by the gradient accumulation bug~\cite{lysandre-fixing}.} of 80 contexts, resulting in 154K tokens per a mini-batch.
We leverage DeepSpeed ZeRO-3\footnote{https://github.com/microsoft/DeepSpeed} and train using 24 H100 GPUs.

For SFT, we employ the same hyperparameter training configuration as in the CPT, except for changes to the context length, which is set to 4,096, and the effective batch size, which was set to 256.
We set up an autoregressive training objective and trained by masking everything except for the answer part of the prompt.
Each sample is padded to match the length of the sample with the maximum sequence in the batch. 
Samples exceeding the maximum sequence length of 4,096 tokens are excluded. The chat template is used for prompting. 

\subsection{Evaluation Dataset}
\label{sec:evaluation-mcq}
Since the original dataset is small, we generate additional variations by applying answer choice permutations. Specifically, as each question in our test set has four answer choices, we systematically permute the order of these choices to create new variants of the same question. For each question, there are $4! = 24$ possible permutations of the answer choices. For every permutation, we update both the answer options and the corresponding correct answer label accordingly, ensuring that the content and logic of the question remain consistent. This augmentation strategy significantly expands the dataset size, enhances model robustness to variations in answer ordering, and helps prevent overfitting to specific patterns in the original dataset.
\begin{table}[!t]
    \caption{The number of MCQs in the test set for each JP1 examination level for evaluations}
    \centering
    \label{tab:mcq-split}
    \begin{tabular}{l|l|l|l}
    \toprule
        Level & Engineer & Professional & Consultant \\\midrule
        Test set & 10 & 12 & 20 \\
        Test set after perm. & 240 & 288 & 480 \\
    \bottomrule
    \end{tabular}
\end{table}

\subsection{RAG Conditions}
The embedding vectors for RAG are built using a chunk size of 380 tokens without overlapping. During retrieval, we retrieve the top 20 most relevant chunks based on similarity search. This setup ensures that the model has access to domain-specific knowledge while maintaining computational efficiency.

\subsection{Experiment Conditions}
We conduct experiments under the following conditions. 1) To isolate the effect of agentic prompting, we compare two inference regimes: \textbf{standard prompting} and \textbf{agentic prompting}. 2) Because the order in which examples appear during supervised fine-tuning (SFT) can bias the final model, we also probe for a “curriculum” effect by training under two data-ordering schedules. Shuffled Data: All training data is randomly shuffled before fine-tuning. And CoT Last: CoT samples are presented at the final stage of SFT to assess their influence on model reasoning. 3) The number of iterations is set based on our experimental configuration: \#Attempts = 5, \#Cycles = 11, setting the limit for the number of reasoning cycles before generating a final answer.
4) Additionally, we evaluate the performance of GPT-4 (gpt-4-0618) as an external baseline. To evaluate the effectiveness of our model, we use accuracy rate, which measures the correctness of the model’s outputs compared to the expected answers. 

\begin{table*}[t]
  \centering
  \sisetup{table-format=2.0}      
  \resizebox{\textwidth}{!}{%
  \begin{tabular}{
      @{}l c cccc *{6}{S[table-format=2.0]}@{}}
    \toprule
    \multirow{2}{*}{\textbf{Model}} &
    \multirow{2}{*}{\textbf{CPT}}  &
    \multicolumn{4}{c}{\textbf{SFT}} &
    \multicolumn{2}{c}{\textbf{Engineer}} &
    \multicolumn{2}{c}{\textbf{Professional}} &
    \multicolumn{2}{c}{\textbf{Consultant}} \\
    \cmidrule(lr){3-6}\cmidrule(lr){7-8}\cmidrule(lr){9-10}\cmidrule(lr){11-12}
     & & Shuffle & Public & JP1 ReAct & JP1 CoT
       & {Standard} & {Agentic} & {Standard} & {Agentic} & {Standard} & {Agentic} \\
    \midrule
    1 & \xmark & \xmark & \xmark & \xmark & \xmark & 68 & 71 & 35 & 35 & 42 & 46 \\
    2 & \cmark & \xmark & \xmark & \xmark & \xmark & 78 & 68 & 46 & 28 & 52 & 42 \\
    3 & \cmark & \cmark & \cmark & \cmark & \xmark &
        65 & \textbf{84} &
        46 & \textcolor{red}{\bfseries 47} &
        45 & \textcolor{red}{\bfseries 56} \\
    4 & \cmark & \cmark & \cmark & \cmark & \cmark & 73 & 74 & 33 & 30 & 50 & 39 \\ 
    5 & \cmark & {\xmark (CoT last)} & \cmark & \cmark & \cmark &
        70 & 82 & 39 & 32 & 49 & 51 \\
    \midrule
    \multicolumn{6}{l}{\textit{GPT-4}} &
        71 & \textcolor{red}{\bfseries 89} & 29 & 40 & 38 & 49 \\
    \bottomrule
  \end{tabular}}
  \caption{Performance results (\%). “Standard” = standard prompting. “Agentic” = Agentic prompting.}
  \label{tab:results}
\end{table*}


\section{Results}
The experimental results, as summarized in Table~\ref{tab:results}, demonstrate that Model 3, which employs the proposed agent fine-tuning methodology, achieves the best overall performance across all evaluation categories. Performance improves when the format of the final SFT stage matches the format used at inference time. Moreover, prompts written in an agentic style attain a higher ceiling of performance than standard prompts.
Incorporating CPT yields substantial improvements in performance, particularly at the Engineer and Professional levels, evidencing the benefit of leveraging domain-specific knowledge.

In contrast, adding CoT trajectories during fine-tuning does not improve performance; instead, it leads to a noticeable decrease, particularly at higher expertise levels (Professional and Consultant). This effect may be attributed to the fact that CoT trajectories actually reduce the number of reasoning iterations the model undertakes. As a result, the model may not fully leverage its iterative reasoning capabilities, which appear to be especially important for more complex, professional-level tasks. This suggests that encouraging more extensive reasoning iterations—rather than constraining them via CoT trajectories—can be more beneficial for achieving high performance on advanced tasks.
Overall, these findings validate the effectiveness of our agent fine-tuning and prompting approach, demonstrating that it not only surpasses previous baselines but also outperforms GPT-4 on advanced-level examinations, including Professional and Consultant certifications.

\section{Conclusion and future work}
In this paper, we propose an efficient approach to fine-tuning LLMs for autonomous reasoning and decision-making within Hitachi's JP1 middleware. Our method leverages domain-specific datasets, including manuals and agent trajectories, and incorporates CPT prior to SFT to enable precise adaptation. Additionally, we systematically explore the role and positioning of CoT trajectories, finding that excluding CoT yields the best performance for our fine-tuned LLM. This leads to significant performance improvements on JP1 certification exam tasks—achieving gains of up to 16\% over the base model. These results demonstrate enhanced domain-specific reasoning capabilities and provide new insights into the fine-tuning process for domain-specific applications.

Future work will focus on optimizing model performance and efficiency through task-specific fine-tuning strategies and systematic evaluation of data quality and diversity. By tailoring fine-tuning for specific JP1 applications and ensuring robust, generalizable training data, we aim to further enhance the cost-effectiveness, scalability, and applicability of LLMs for domain-specific reasoning and decision-making tasks. Beyond the JP1 middleware context, we also see potential applications of agent fine-tuning in broader domains where structured decision-making and reliability are critical, such as economics and finance.

\bibliographystyle{IEEEtran}

\begin{CJK}{UTF8}{min}
\bibliography{main}   
\end{CJK}

\end{document}